\documentclass{article}
\usepackage[utf8]{inputenc}
\usepackage{csquotes}
\usepackage{hyperref}
\hypersetup{
    colorlinks= true,
    linkcolor= cyan,
    filecolor= magenta,      
    urlcolor= cyan,
}
\usepackage{graphicx}
\usepackage{geometry}
\usepackage{subcaption}
\usepackage{amsmath,amssymb,amsfonts}    
\usepackage{algorithm}
\usepackage{algorithmic}
\usepackage{libertine}
\begin{document}


\begin{table}[ht]
\begin{tabular}{p{5.5cm}p{5.5cm}p{5.5cm}}
\multicolumn{3}{c}{\huge \textbf{FLaPS: Federated Learning and Privately Scaling}}                                                                      \\
&&\\
&&\\
&&\\
\large \textbf{Sudipta Paul  }                                        & \large  \textbf{Poushali Sengupta }                                 &\large  \textbf{Subhankar Mishra }                                     \\
&&\\
School of Computer Sciences                           & Department of Statistics                           & School of Computer Sciences                           \\
National Institute of Science, Education and Research & University of Kalyani                              & National Institute of Science, Education and Research \\
Bhubaneswar, India -752050                            & Kalyani, West Bengal                               & Bhubaneswar, India -752050                            \\
Homi Bhabha National Institute,                       & India - 741235                                     & Homi Bhabha National Institute,                       \\
Anushaktinagar, Mumbai - 400094, India                & \multicolumn{1}{l}{}                               & Anushaktinagar, Mumbai - 400094, India                \\
sudiptapaulvixx@niser.ac.in                           & \multicolumn{1}{l}{tua.poushalisengupta@gmail.com} & smishra@niser.ac.in                                   \\
ORCID: 0000-0001-6561-997X                            & \multicolumn{1}{l}{ORCID: 0000-0002-6974-5794}     & ORCID: 0000-0002-9910-7291                           
\end{tabular}
\end{table}

\begin{abstract}

Federated learning (FL) is a distributed learning process where the model (weights and checkpoints) is transferred to the devices that posses data rather than the classical way of transferring and aggregating the data centrally. In this way, sensitive data does not leave the user devices. FL uses the FedAvg algorithm, which is trained in the iterative model averaging way, on the non-iid and unbalanced distributed data, without depending on the data quantity. Some issues with the FL are, 1) no scalability, as the model is iteratively trained over all the devices, which amplifies with device drops; 2) security and privacy trade-off of the learning process still not robust enough and 3) overall communication efficiency and the cost are higher. To mitigate these challenges we present Federated Learning and Privately Scaling (FLaPS)\footnote{This research was partially supported by Department of Science and Technology Grant.
\par Grant ID:NRDMS/UG/S.Mishra/Odisha/E-01/2018.\par This paper has been accepted at \href{https://drpskfec.github.io/slice2020/call-for-papers.html}{SLICE 2020}} architecture, which improves scalability as well as the security and privacy of the system. The devices are grouped into clusters which further gives better privacy scaled turn around time to finish a round of training. Therefore, even if a device gets dropped in the middle of training, the whole process can be started again after a definite amount of time. The data and model both are communicated using differentially private reports with iterative shuffling which provides a better privacy-utility trade-off. We evaluated FLaPS on MNIST, CIFAR10, and TINY-IMAGENET-200 dataset using various CNN models. Experimental results prove FLaPS to be an improved, time and privacy scaled environment having better and comparable after-learning-parameters with respect to the central and FL models.


\end{abstract}


\section{Introduction}\label{intro}


Smartphones, Tablets, e-reader devices etc. have become an intrinsic part of our daily life with its different sophisticated features. According to a survey\cite{b22} the vast majority of Americans – 96\% has an ownership of cellphone. Since 2011, the share of smartphone users in America increased from 35\% to 81\%. In India, it is evaluated that by 2022 the smartphone users will extend to 442 million\cite{b19}, also the spiking rate of smartphone users in India was predicted to get more than 28 percent by 2018\cite{b20}. These \enquote{on the go} devices are frequently carried to a lot of places. The powerful sensors that these devices consist of i.e. camera, GPS, microphone etc. have the power to generate and preserve data from different places whether it is \enquote{on the move with steady connection} or \enquote{idle and not under steady connection}. From storing contact numbers, songs to important documents, doing banking transaction, insurance payment, taking and storing photos as a token of remembrance of the personal milestones etc.- these smart goodies have become an identity to its owner and have an access to an extraordinary amount of private and sensitive information about the real world. These data have a huge variation, distribution and are non-iid and sensitive to privacy attacks such as: DDOS, man in the middle attack, SQL injection, eavesdropping attack etc. These private and sensitive raw data are very useful for the improvement of different machine learning models\cite{b1} i.e. image processing models, speech to text generation, mobile keyboard spelling correction function for different languages, face recognition function etc.
\par  \textbf{Challenges} The aforementioned machine learning models are normally trained in a centralized environment. But the sensitive nature of these smart devices' data is strongly against to store them centrally. Storing at a central storage include uploading them to a cloud based or hardware based central server. Even if the transaction is safe, there are possibilities that the information can be leaked at the time of the training too. Thus we need a robust privacy protection system, which can be scaled in every steps of the learning reasonably.

\par \textbf{Distributed Learning} To solve these challenges, at first \textit{Distribution Learning}\cite{b21}(DL) was proposed. In their 2017 work\cite{b25} Diakonikolas et al. focused on non parametric density estimation of $l_1$ and $l_2$ distribution on multiple servers' data which gave the upper and lower bounds on the communication complexity of DL. In the same year\cite{b25}, Huang et al. proposed a DP based ADMM(Alternating directed methods of multipliers) system to prevent the security leakage in DL with lower utility. In their 2019 survey\cite{b24} Ben-nun et al. gave an analysis on how the DL has been evolved through time with: Architectural, Data, Algorithmic and Programming parallelism. The setbacks of DL observable from these works are: it is more focused on the power saving using parallel computing, the data is assumed to be iid - with almost same size at each server, the number of servers to be trained is limited, and all the servers are connected. Therefore to meet our aim, we need DL but in elevated and different shape.

\par \textbf{Federated Learning} In their 2017's work McMahan et al. first proposed a system to acknowledge all these problems which is known as \textit{\enquote{Federated Learning}}. In this learning technology, the model itself goes to the data rather than data comes to be stored and trained in a central manner. This comparatively newer technology is one of the biggest research field right now with its enormous possibilities of contributions towards different fields. Google has already used FL commercially in several applications i.e. Hard et al. 2019's work\cite{b27} on next word prediction and Chen et al.'s work\cite{b28} on out-of-vocabulary-word in the same year on the Google keyboard.
\par  \textbf{Federated Learning - Drawbacks} We will know more about how FL works in the Section \ref{background}. But there are some definite setbacks of FL that require more research, i.e.:
(1) Device drop, (2) Privacy traded off with the utility due to the lack of scalability in the usage of DP. Zhu et al.\cite{b14} and Caldas et al.\cite{b9} tried to give a solution by acknowledging the heavy hitters and probabilistic quantization on user devices respectively to mitigate the device drop situation. But they were all for lossy compression of data, which is actually not good for the data utility. Also no scalability mechanism has ever been pushed to the system of FL to prevent the utility-privacy trade-off while aggregating DP reports. 
 \par \textbf{Contributions} Acknowledging these challenges we are proposing a distributed FL system, \textbf{FLaPS} using a novel privacy scaling method in four steps to elevate the normal FL. The contributions of our model are:
 \begin{itemize}
     \item  It is in the middle of DL and FL.
     \item No two devices are connected among themselves but to the cluster centres only; 
      \item The trained model quality is better or comparable with that of normal FL;
      \item Two privacy scaling steps at the cluster centres and two in the central server, gives novel privacy scaling mechanism. 
      \item Number of communication channels are scalable in FLaPS. The communication is scalable here by choosing the number of cluster heads only, e.g. normal FL needs to visit 2 million devices, but FLaPS needs to visit 100 - 2000 cluster centres only.
      \item The clustering approach let the central server inspect the device drop situation after a definite time intervals economically. 
       
      \end{itemize}
 
 \par The rest of the paper is arranged in the following way:
 Following by the Introduction in Section \ref{intro}, Section \ref{rw} describes the related works in this field, Section \ref{background} describes the necessary background to build the  base to understand our experiment, Section \ref{algorithm} describes our proposed algorithm, Section \ref{exp} describes our experiment in details, Section \ref{ra} discusses the result and analysis, and lastly Section \ref{conclusion} discusses the future scope.

\section{Related Work}\label{rw}

The main challenges according to Kairouz et al.\cite{b15} that FL faces in the time of the whole process are: 
\begin{itemize}
    \item The data is massively distributed over a large number of learning nodes.
    \item The data at nodes may be extracted from different distribution which may not be same as the one at the central server. This phenomenon is defined as non-IID. This affects the training quality of the transferred model a lot.
    \item The nodes may hold different number of examples to train the models which may be varied from very less to very huge in size, which is highly unbalanced.
    \item The devices might get dropped at any time of the whole process without any prior notice.
    \item In need of a security system which actually ensures the security of the whole aggregated system. 
\end{itemize}
\par Two main aims of FL are maintaining the communication efficiency even if the network connection is not steady and maintaining the privacy and security of the sensitive data holders at the time of the model training. By communication efficiency we are aiming the situations where even after the device getting dropped, the turn around time will be lesser to go back again and after the updation of weights in the central server, the model quality is still good. 
\par Various solutions has been discussed over time acknowledging them. Bonawitz et al. in their 2016 work\cite{b8} proposed a communication efficient secured aggregation protocol based on secret sharing,   Kone$\check{c}n\acute{y}$ et al. in their 2017 work\cite{b2} proposed two solutions e.g. \textit{structured updates:} where the model learns an update from a restricted space which is later parametrized with a smaller number of variables and sketched updates: where the model learns the whole update, then compress and send it to the server, for better communication efficiency.
\par In their 2019 work\cite{b9} Caldas et al. initiates two new schemes to lower the communication costs: (1) they used lossy-compression on the global model at the time of transaction it from the server to the clients; and (2) a new technique, Federated Dropout, which enabled the users to train the global model efficiently at the local setup, using compact subsets of the model, which in turn gave a cutting in the communication and local computation cost. 
\par In their 2019 work\cite{b10} Bonawitz et al. presented a new model for secure and efficient communication. The proposed model auto-tune the aggregation, using particular characteristics of random rotation e.g. the foreseeable distribution of vectors after-rotation and the basic modular wrapping in secure aggregation. In their work\cite{b11} in 2019, Jiang et al. proposed that a model trained using a standard datacentre optimization method is much harder to personalize, compared to one trained using \textbf{FedAvg}.
\par In their 2019 work\cite{b12}, Sun et al. observed that in the absence of defences, the performance of the attacks largely depends on the fraction of adversaries present and the “complexity” of the targeted task. Moreover, they showed that \textit{norm clipping and “weak” DP mitigate the attacks without hurting the overall performance}. In 2020, Reddi et al.\cite{b13}'s results give an insight on the interaction between client diversity and efficiency of communication . They plotted the federated varieties of adjustable optimizers, like ADAGRAD, ADAM, and YOGI. They inspect their convergence in the presence of diversified data for common nonconvex environment.  
\par In their 2020 work, Zhu et al.\cite{b14}, based on big wheels in a sample of user-produced data flows, propose a distributed and privacy-preserving algorithm. They grasp the sampling and threshold properties of their  algorithm to demonstrate that it is intrinsically obeys the rules of DP, without called for extra noise. 
\par None of these solutions ever highlighted the following things:
\begin{itemize}
    \item The turn-around time bottleneck for each round when a device gets dropped.
    \item Utility and privacy, that are maintained through each step of the security maintenance mechanism.
\end{itemize}
These two shortcomings have mainly fueled us to propose the solution in the Section \ref{algorithm}.



\section{Background Theories and Inspiration}\label{background}


In this section we will discuss about the necessary background of DP\cite{b5} techniques that we used at the time of the experiments and FL. 

\subsection{Federated Learning}
The term \enquote{\textit{Federated Learning}} or \textbf{FL} was first coined by McMahan et al.\cite{b1}. By their    words - \textit{\enquote{We term our approach FL, since the learning task is solved by a loose federation of participating devices (which we refer to as clients) which are coordinated by a central server.}}

\par The aim of FL is to learn a shared model by local aggregation. In this system the data is left in the devices and the central server learns the aggregation of the local weight updates after the training of the shared model through FedAvg algorithm. FedAvg combines local stochastic gradient descent (SGD) on each client at a central server that performs the weight averaging. The general objective is of the form:
\begin{equation}
    \min_{w \in \mathbf{R}^d} f(w) \text{ where } F_k(w) = \frac{1}{n_k} \sum_{u \in P_k} f_i(w)
\end{equation}

where, we typically take $f_i(w) = l(x_i, y_i; w)$, that is, the loss of the prediction on example $(x_i, y_i)$ made with model parameters $w$.

\par FL didn't take the stage in one day. A sequence of a lot of other works paved the way to mark its entry as a variation of \enquote*{\textit{Distributed Learning}}. In their work in 2017\cite{b4} Bonawitz et al. showed how the FL works. The main participants are devices and FL servers. The FL servers are cloud-based and distributed. When the devices report to the FL server that they are set, then a problem in shape of \textit{FL population} is then announced to the devices. The problems are mainly \textit{learning problems.} After establishing the connection with devices the server then sends a \textit{FL plan}, consists of a TensorFlow graph, and the programs and checkpoints of how to execute it. This establishes a round among the devices and the central server. Then the server sends the global parameters and \textit{FL checkpoints} to the devices for the model to be trained. When the devices finish the model training they send back the updated \textit{Fl Checkpoints} back to the central server. 
\par Summarizing the flow of FL: Devices let know the FL server that they are ready, the not ready ones are told to participate later $\longrightarrow$
 Central Server studies the model checkpoints from  storage $\longrightarrow$
     Model and checkpoints are sent to the ready devices $\longrightarrow$
     On-device training takes place, model update is reported back $\longrightarrow$
     Server aggregates updates using FedAvg into the global model $\longrightarrow$
     Server updates the global model checkpoints into the storage.

\subsection{Differential Privacy(DP)}
To satisfy individual privacy in a sensitive, statistical database, DP\cite{b5} has now become the standard. Here, we use two DP algorithms. The description of the algorithms are following: 
\subsubsection{\textit{BUDS}}
BUDS\cite{b17} or \textit{Balancing utility and Differential privacy by shuffling} algorithm is divided into two main parts. Attribute reduction and iterative shuffling. Let's have a dataset containing $n$ number of rows and $k$ number of attributes. In the first part of this mechanism, a query function is used to know about the necessary attributes to answer that particular query. These attributes tied up together and let, the reduced number of attributes becomes $m$. In the second part, let there be $S$ number of shufflers to shuffle the data. Before shuffling, the $m$ numbers of attributes are divided into $S$ groups and the whole dataset is divided into $1,2, ..., t$ batches containing $n_1, n_2, ...., n_t$ number of rows  respectively where $n_1 \simeq n_2 \simeq ..... \simeq n_t$. Now, each group of attributes from each batch chooses a shuffler from $S$ number of shufflers randomly without replacement technique and go for independent shuffling. This process is iterated until the last batch goes for shuffling. After iterative shuffling, the final report is generated.

\subsubsection{\textit{ARA}}
ARA\cite{b16} or \textit{Aggregated RAPPOR and Analysis} is a centralized DP approach. In this system, the RAPPOR \cite{b32} reports were taken. Then a weighted sum is calculated using the ARA constants according to the positions on the report bit string. The constants have been identified before hand, using TF-IDF (term frequency - inverse document frequency) on the RAPPOR reports, with continuous sampling. The details of this approach can be found here in the Paul et al's \cite{b16} work. 
The aggregated approach of ARA gave a satisfactory result which is a direct inspiration to use it at the central server and cluster centers to retrieve the weights and data respectively.
\subsection{Clustering Algorithm}
Cluster analysis is a technique by which a set of data-points, objects or entities gets divided into groups in such a way that those data-points, objects or entities are similar among themselves in some senses. We use the Hartigan \& Wong\cite{b6} variation of \textit{k-means clustering} algorithm.  Suppose we have $n$ data points in $d$-dimensional space $R^d$ and an integer $k$. Here, The aim is to determine $k$ centres in $R^d$ such that the mean squared distance among the $n$ data points with its centres is minimum. 


\section{FLaPS: Federated Learning and Privately Scaling}\label{algorithm}
As we have learned the background theories, now we are describing our system in the following:
\begin{itemize}
    \item Suppose, we have $n$ numbers of user devices who agreed to be trained. Let the central server choose a number $k$ ,where $2 < k < n$. All the $n$ users will get clustered using \textit{k-means clustering} algorithm depending on the budget of $k$. After this the $k$ cluster centres will only have communication with the central server. 
    \item The $k$ cluster centres then send a message to the other cluster members to collect the differentially private data (produced using \textit{BUDS} algorithm). The collected data will be then aggregated (using \textit{ARA} algorithm).
    \item The $k$ cluster centres will then train the already downloaded model from the central server with the aggregated data. 
    \item After training, the weights and checkpoints of the trained model at each cluster centres will again get differentially privatized using the \textit{BUDS} algorithm.
    \item The reports are then uploaded to the central server and the \textit{ARA} algorithm then aggregates them to get a final report and update the central model according to the final report and the \textit{FedAvg} algorithm.
\end{itemize}

\begin{figure*}
     \centering
     \begin{subfigure}[b]{0.32\textwidth}
         \centering
         \includegraphics[width=\textwidth]{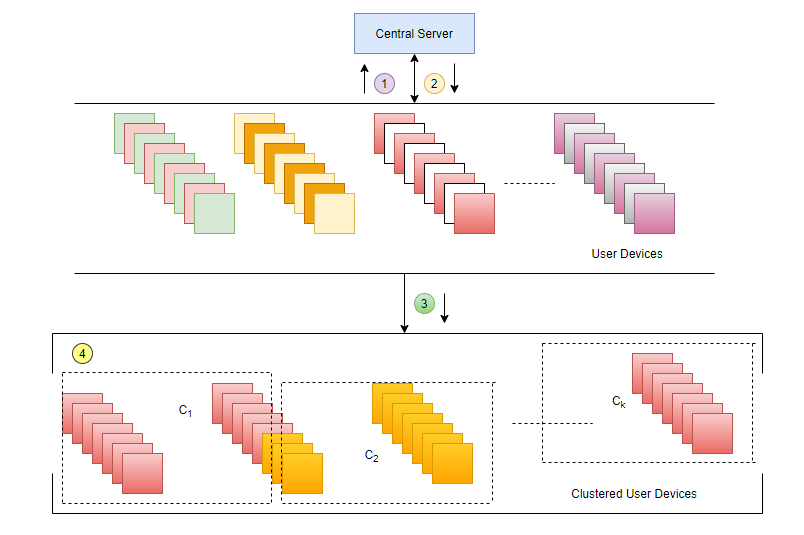}
         \caption{At the server }
         \label{fig:y equals x}
     \end{subfigure}
     \hfill
     \begin{subfigure}[b]{0.33\textwidth}
         \centering
         \includegraphics[width=\textwidth]{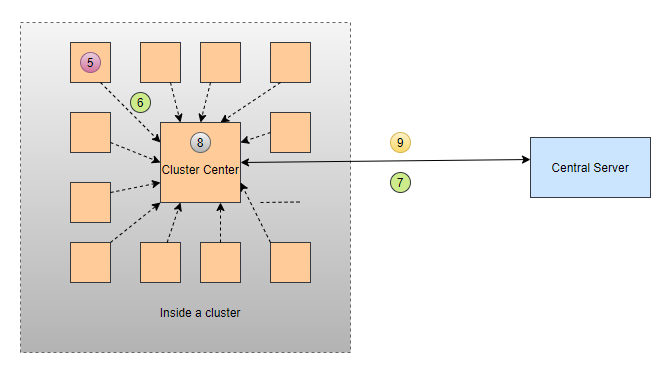}
         \caption{At each $k$ clusters}
         \label{fig:three sin x}
     \end{subfigure}
     \hfill
     \begin{subfigure}[b]{0.32\textwidth}
         \centering
         \includegraphics[width=\textwidth]{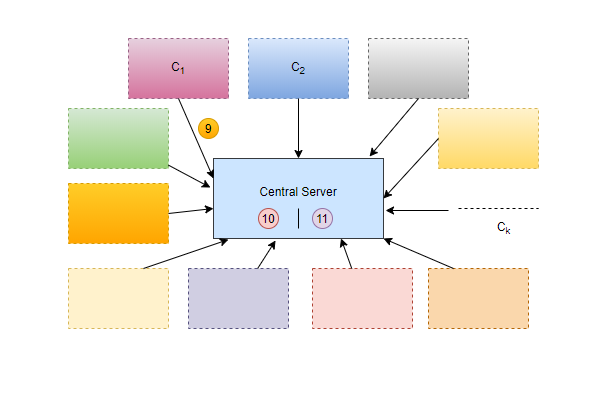}
         \caption{At the server}
         \label{fig:five over x}
     \end{subfigure}
        \caption{The whole flow of FLaPS: (1) The User devices let know the Central Server that they are ready (2) The Central Server randomly choice a budget for k and send it to the users (3) The budget of $k$ is broadcasted to the devices. (4) $k$ Clusters formed with the help of the k-means algorithm.(5) Using BUDS algorithm the user data is differentially privatizing. (6) The DP report from each user is collecting at the cluster centre. (7) The training model is downloaded from the central serve.r (8) The model is trained using the ARA aggregated user report. (9) The weights of the model produces a DP report and is uploaded to the central server.(10) The reports are aggregated using ARA. (11) The weights are updated using FedAvg Algorithm}
        \label{flow}
\end{figure*}

The formal algorithm of the above description is given in Algorithm \ref{algo}. The flow chart of the proposed algorithm's architecture is given in the Figure \ref{flow}
\begin{algorithm}[htbp]

\caption{\textbf{FLaPS: Federated Learning and Privately Scaling} \\ \textbf{Input}: $n$ user devices ready to be trained with their    data
\\ \textbf{Output}: A centrally federated trained machine learning model }
\begin{algorithmic}
\STATE \textbf{At the Server:}
\STATE $n$ idle users is selected
\STATE $k \in n$ gets randomly selected
 \FOR{each $k \in n$} 
        \item $kmeans\_clustering(k,n)$
         \ENDFOR

\STATE \textbf{At Each $k$ Clusters:}
\STATE $n$ is the set of user devices
\STATE $C = {C_1, C_2, C_3, ..., C_k}$ is the set of clusters where $k \geq 2$
\FOR{each user $n_j \in C_i$} 
    \item $Report_j = BUDS(a,n_j)$, where $a$ are attributes
    \item $Report_j$ is transmitted to the $C_i$ centres
    \ENDFOR
\FOR{each $C_i \in C$, where each $C_i$ is the server centres}    
    \item Model downloaded from the central server
    \item $W_{agg} = ARA(Report_j, Reports)$
    \item Model gets trained with $W_{agg}$
    \item $Report_k = BUDS(w,C_k)$, where $w$ is the weights of the trained model
    \item $Report_k$ is transmitted to the central server
\ENDFOR

\STATE \textbf{At the Server:}
 \FOR{each $Report_k \in Reports$} 
        \item $W = ARA(Report_k, Reports)$
        \ENDFOR
        \STATE $W_{fed} = FedAvg(W)$
        \STATE Testing Accuracy, Prediction Accuracy, AUC, F1 Score is calculated
         
\end{algorithmic}
\label{algo}
\end{algorithm}


\section{Experiment}\label{exp}
\subsection{System Description}
The system specification is - UBUNTU 19.04 64-bit Kernel Linux 5.0.0-37-generic x86\_64 MATE 1.20.4 OS with 502.6 GB of memory and Intel Xeon(R) Gold 6138 CPU @ 2.00Ghz x 80, 4 GPU components(GeForce RTX 2080i) with CUDA version 10.1, driver version 430.50 with 6 TB external disk space. The R-studio with R-version 3.6.1, Keras(version 2.2.5.0) and Tensorflow(version 2.0.0) has been used, with a steady internet connection with a 8.70 Mbps download speed and a 11.96 Mbps upload speed.


\begin{table}[hb]
\centering
\caption{Description of datasets}
\label{dataset}
\begin{tabular}{llrrrrr}
\hline
Dataset Name      & Image Scale & Train   & Test   & Valid-ation & Image Size & Class \\
\hline
\noalign{\vskip 0.25 cm}
MNIST             & Grey        & 60000   & 10000  & Null       & 28x28      & 10    \\
\noalign{\vskip 0.1 cm}
CIFAR-10          & RGB         & 50000   & 10000  & Null       & 32x32      & 10    \\
\noalign{\vskip 0.1 cm}
Tiny-ImageNet-200 & RGB         & 100,000 & 10,000 & 10,000     & 64x64      & 200 \\
\hline
\end{tabular}
\end{table}
\subsection{Learning Model and Data Description}
The description of the datasets are given in the Table \ref{dataset}. For thrice of the datasets pre-trained CNN models has been used. The layer to layer description is being given in the following Tables \ref{model_cifar10}, \ref{model_mnist}, \ref{IR} and \ref{mv}.

\begin{table}[htbp]
\caption{Learning model for CIFAR10 data}
\begin{center}
\begin{tabular}{llr}
\hline
\textbf{Layer(type)}&\textbf{Output shape}&\textbf{Param \# } \\
\hline 
conv2d (Conv2D) & (None, 32, 32, 32)  & 896\\

activation (Activation) &(None, 32, 32, 32)& 0 \\

conv2d\_1 (Conv2D) &(None, 30, 30, 32) &9248 \\

activation\_1 (Activation) &(None, 30, 30, 32) & 0              \\

max\_pooling2d (MaxPooling2D)&(None, 15, 15, 32)& 0  \\

dropout (Dropout)   &(None, 15, 15, 32) &0    \\

conv2d\_2 (Conv2D) & (None, 15, 15, 32)& 9248  \\

activation\_2 (Activation)  & (None, 15, 15, 32) & 0 \\

conv2d\_3 (Conv2D) & (None, 13, 13, 32) &9248 \\

activation\_3 (Activation) &(None, 13, 13, 32) &0    \\

max\_pooling2d\_1 (MaxPooling2D)& (None, 6, 6, 32) &0   \\

dropout\_1 (Dropout)&(None, 6, 6, 32)   &0 \\

flatten (Flatten)  &(None, 1152)&0 \\

dense (Dense) &(None, 512) & 590336 \\

activation\_4 (Activation) &(None, 512)& 0 \\

dropout\_2 (Dropout)&(None, 512)&0 \\ 

dense\_1 (Dense)&(None, 10)&  5130   \\

activation\_5 (Activation)  &(None, 10) &0\\
\hline
Total params: &&624,106\\
Trainable params: &&624,106\\
Non-trainable params:&& 0\\
\hline
\end{tabular}
\label{model_cifar10}
\end{center}
\end{table}

\begin{table}[htbp]
\caption{Learning model for MNIST data}
\begin{center}
\begin{tabular}{llr}
\hline
\textbf{Layer(type)}&\textbf{Output shape}&\textbf{Param \# } \\
\hline 
conv2d (Conv2D) & (None, 26, 26, 32)  & 320\\

conv2d\_1 (Conv2D) &(None, 24, 24, 64) &18496 \\

max\_pooling2d (MaxPooling2D)&(None, 12, 12, 64)& 0  \\

dropout (Dropout)   &(None, 12, 12, 64) &0    \\

flatten (Flatten)  &(None, 9216)&0 \\

dense (Dense) &(None, 128) & 1179776 \\

dropout\_1 (Dropout)&(None, 128)&0 \\ 

dense\_1 (Dense)&(None, 10)&  1290   \\
\hline
Total params: &&1,199,882\\
Trainable params: &&1,199,882\\
Non-trainable params:&& 0\\
\hline
\end{tabular}
\label{model_mnist}
\end{center}
\end{table}

\begin{table}[htbp]
\caption{InceptionResNetV2\cite{b31} model for Tiny-Imagenet-200 data}
\label{IR}
\begin{center}
\begin{tabular}{llr}
\hline
Layer (type)                  & Output Shape       & Param \#   \\
                             
                              \hline
inception\_resnet\_v2 (Model) & (None, 5, 5, 1536) & 54336736   \\
                             
dropout\_2 (Dropout)          & (None, 5, 5, 1536) & 0          \\
                              
flatten\_1 (Flatten)          & (None, 38400)      & 0          \\
                             
dropout\_3 (Dropout)          & (None, 38400)      & 0          \\
                              
dense\_1 (Dense)              & (None, 200)        & 7680200    \\
                              
                              \hline
Total params:                 &                    & 62,016,936 \\
Trainable params:             &                    & 7,680,200  \\
Non-trainable params:         &                    & 54,336,736\\
\hline
\end{tabular}
\end{center}
\end{table}

\begin{table}[htbp]
\caption{MobileNetV2 model\cite{b30} for Tiny-Imagenet-200 data}
\label{mv}
\begin{center}
\begin{tabular}{llr}
\hline
Layer (type)                   & Output Shape       & Param \#   \\
\hline
mobilenetv2\_1.00\_224 (Model) & (None, 7, 7, 1280) & 2257984    \\
dropout\_4 (Dropout)           & (None, 7, 7, 1280) & 0          \\
flatten\_2 (Flatten)           & (None, 62720)      & 0          \\
dropout\_5 (Dropout)           & (None, 62720)      & 0          \\
dense\_2 (Dense)               & (None, 200)        & 12544200   \\
\hline
Total params:                  &                    & 14,802,184 \\
Trainable params:              &                    & 12,544,200 \\
Non-trainable params:          &                    & 2,257,984\\
\hline
\end{tabular}
\end{center}
\end{table}
\subsection{Experiment Description}


The experiment can be easily divided into re-usable sub-components to write the program efficiently, i.e. central server, communication channels, clients and cluster formation. To impose a secure channel the \textit{BUDS} algorithm has been implemented for generating DP reports. It has been used in two out of four communication steps in the proposed model. Also, the users are connected through a simple socket connection to the central server and the respective cluster centers. Find the codes here: github.com/smlab-niser/flaps. \par The experiment works like this -
\begin{enumerate}
    \item The dataset is divided randomly among the clients. The attributes that each client has to maintain are their user-ID, Number of images they have been allotted, Cluster-ID, and the maximum and minimum Indexes of the allotted images. The value of Cluster-ID gets changed every time according to the central server demand. 
    \item The value of these attributes are one-hot encoded and saved at the individual client.
   \item The central server tries to find which client devices are idle or ready to train a machine learning model at this stage. After finding those client devices the central server then randomly choose the budget of $k$. With respect to the budget of the $k$, the clusters formed among the user devices using the \textit{k-means clustering algorithm}.  \\ For our experimental purpose 200 clients has been chosen and the budget of $k$ is being varied from 2-20.
    \item After the cluster formation happened, the cluster centres of the $k$ clusters send a message to the other clients in the clusters to find the range of the image indexes they possessed individually. The max index and the min index attribute gets together to form a single attribute at each client side and then with the help of a shuffler the values of each attribute get shuffled inside a cluster. Suppose the shuffler has $g$ channels and there are $m$ attributes in total. If there are $n$ attributes, that are responsible for the answer of the query asked then, $g = m - n + 1$. An example is given in the Table \ref{example}. The one hot encoded values are being ignored here for the sake of the better understanding of the readers but one-hot encoding has been used in the actual experiment. After shuffling, statistics of the indexes left as it was but the user-IDs and the number of images per clients get shuffled. Therefore, a differentially private report is being generated here.
    
    \begin{table}[htbp]
    \centering
    \caption{Empirical Scenario of Shuffling}
    \begin{tabular}{c c l|c c l}
    \hline
         &\textbf{Before} & \textbf{Shuffling}&&\textbf{After}&\textbf{Shuffling} \\
         \hline
         \textbf{User}&\textbf{\# images}&\textbf{Max\_index}& \textbf{User}&\textbf{\# images}&\textbf{Max\_index}\\
         \textbf{-ID}&&\textbf{:Min\_index}&\textbf{-ID}&&\textbf{:Min\_index}\\
         \hline
         1 & 200 & 300:501 & 3 & 203  & 600:801\\
         2 & 201 & 600:801 & 1 & 200 & 900:1102\\
         3 & 202 & 900:1102& 2 & 202 & 1:198\\
         4 & 198 & 1:198   & 5 & 199 & 487:686\\
         5 & 199 & 487:686 & 4 & 198 & 300:501\\
         6 & 203 & 1200:1403& 6& 201 & 1200:1403\\
         \hline
    \end{tabular}
    \label{example}
\end{table}
\item After getting the reports through a socket connection, an aggregation step to get full insight from all these reports takes place. The \textit{ARA} architecture has been used here. This aggregation happens at all the $k$ cluster centres for its respective clusters. The $k$ clients then download the model from the central server and train the model with the help of the statistics it gained from the aggregated differentially private reports.
\item After the training step the weights are then extracted and differentially privatized using \textit{BUDS} and the information loss is calculated. When the loss converges to a certain value the report of the weights are then sent to the central server.
\item After getting the reports from all the $k$ clusters the central server then aggregates them using \textit{ARA} with a loss optimization. According to that aggregation, the central server then replaces the weights with the aggregated(normally averaged) weights.






\end{enumerate}

\section{Result and Analysis}\label{ra}

\begin{figure*}
    \centering
    \includegraphics[width =\linewidth]{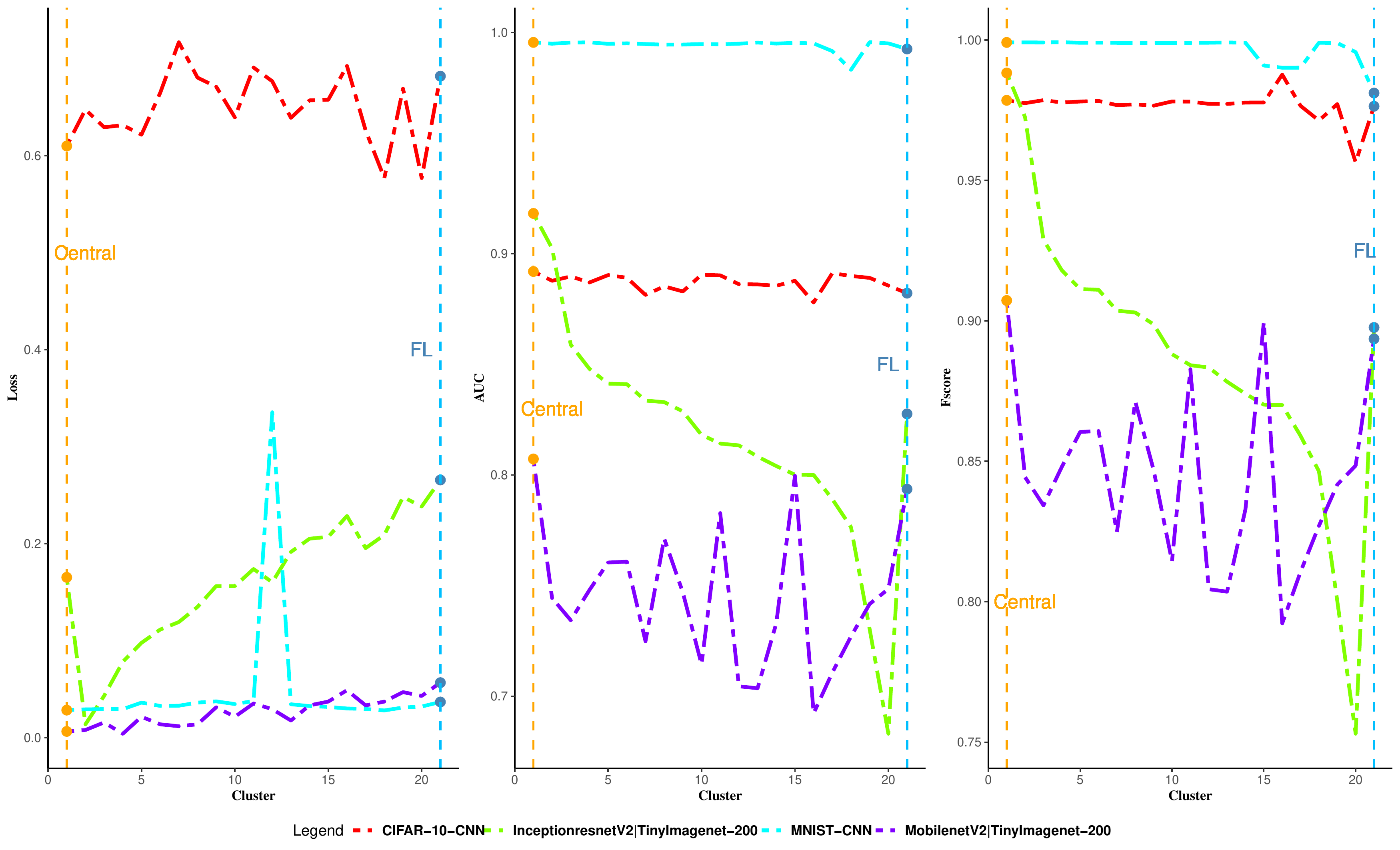}
    \caption{The Loss, AUC and Fscore comparison for all the architectures in FLaPS, Central and FL environment. The orange dot is for the central time and the steel blue dot is for the Federated learning time in the same environment.}
    \label{fig:all-laf}
\end{figure*}

\begin{figure*}
         \centering
    \includegraphics[width = \linewidth]{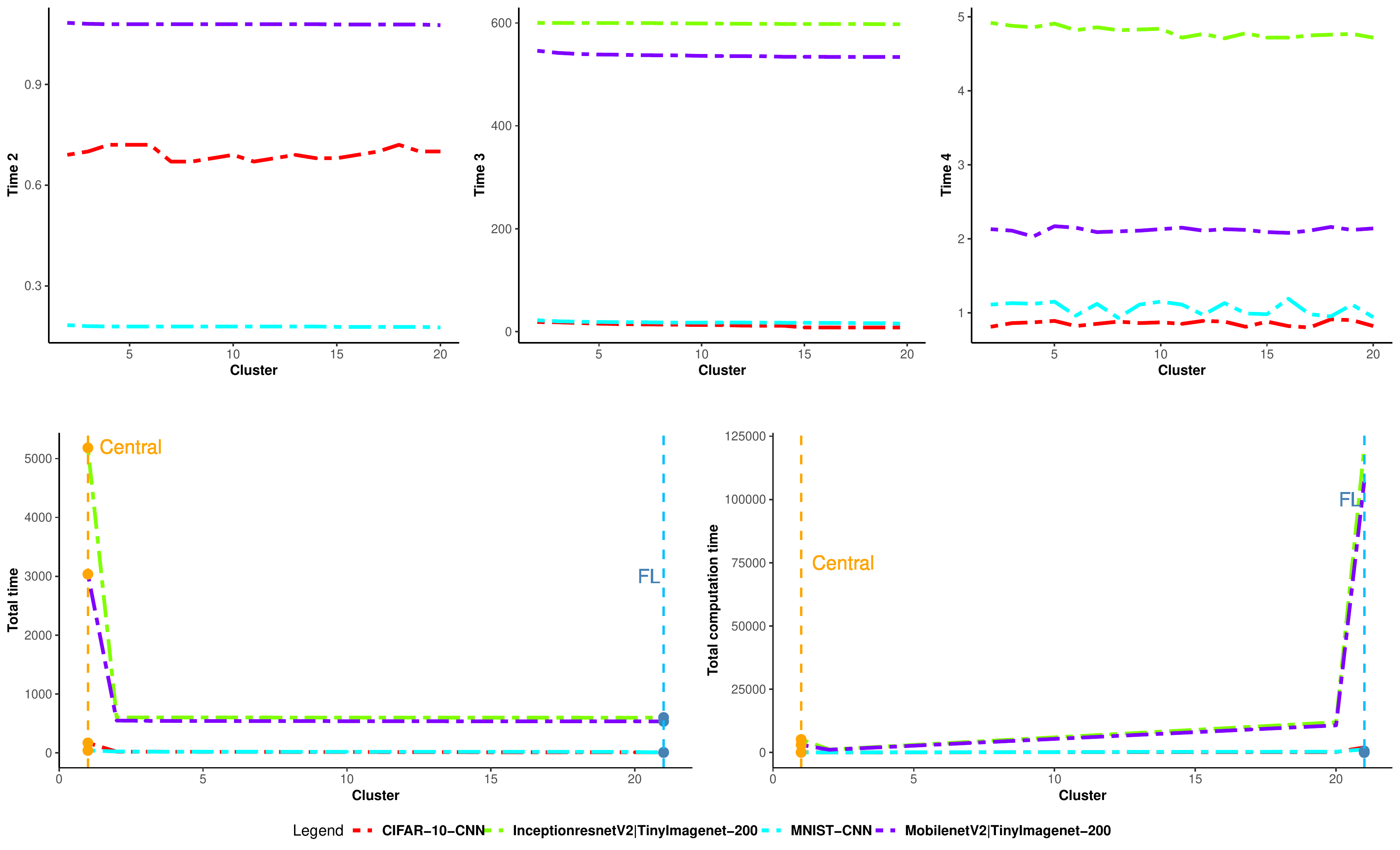}
    \caption{Time comparison across central, FL and FLaPS.  
    }
    \label{fig:all}
    \end{figure*}

      

\subsection{Complexity Analysis}
We have divided our framework into 4 distinct measurable time steps for complexity as well as performance analysis. The time division, it's description and complexity analysis of the proposed FLaPS algorithm for different time division is in Table \ref{complexity}.

\begin{table}[htbp]
\caption{Complexity analysis of FLaPS}
\begin{center}

\begin{tabular}{p{0.1\linewidth} p{0.6\linewidth} p{0.2\linewidth}}
\hline
\textbf{Time}&\textbf{Description}& \textbf{Complexity} \\
\hline 
Time 1 & Random user choice and cluster formation in average time and datasets&$\mathcal{O}(n^2)$\\

Time 2& Generating differentially private report from the clustered users and sending those reports &$\mathcal{O}(n + v^2)$\\
&  to the randomly chosen cluster centres in average time&\\
Time 3 & Average training time per cluster and generating differentially private report of the weights&$\mathcal{O}(dc)$ +Training time per cluster\cite{b18}  \\

Time 4 & Recieving the differentially private reports of weights from each cluster in average plus &$\mathcal{O}(c^2+dc)$\\
& the federated average algorithm run time&\\

\hline
\end{tabular}
\label{complexity}

\end{center}
\end{table}
\subsection{Empirical Analysis}

The results for all the experiments are arranged in the Tables \ref{time} and \ref{laf}. The graphs have also been plotted using these results for further analysis. They are in Figures \ref{fig:all} and \ref{fig:all-laf}. The empirical analysis can also be done from two perspectives: 1) analysing the after training parameters, 2) analysing the time consumption of each and every steps of the algorithm.
\subsubsection{Model quality analysis after training} 
Three parameters: loss, AUC and Fscore are being taken here to analyze the after training model quality. Here, at the time of training: rmsprop is being used as the optimizer function and categorical crossentropy is being used as the loss function. All the result data is given in Table \ref{laf}. 
\par From the loss graph in figure \ref{fig:all-laf} it is evident that the more non-iid and varied the data is, the loss is lesser or equal to the central learning. In case of 2-clustered FLaPS approach, the claim is always true irrespective of architectures and datasets. 


\par AUC measures the quality of the model's predictions irrespective of what classification threshold is chosen. For all the dataset i.e. MNIST, CIFAR-10 and TINY-IMAGENET-200 the proposed clustering approach has produced higher or almost equal AUC in comparison to the FL, which in turn proves that the quality of the learned model is not degrading at all after our training approach. All the result data are in the Table \ref{laf}. From the graphs at Figure \ref{fig:all-laf} it is evident that in case of MNIST-CNN and CIFAR-10-CNN model the Fluctuations of AUC score is almost linear in comparison to the central and the FL models. From the mobilenet and inceptionresnet model AUC score graph it is evident that more the non iid data the model learns better.  
\par The Fscore, is a measure of a test’s accuracy. More the value of Fscore, better the model. The tendency of learning in the Fscore graphs \ref{fig:all-laf} is also looking same as discussed in the previous paragraph regarding the AUC graphs.      
\subsubsection{Time analysis}
For the second approach of analysis the Table \ref{time} will be used. Here, the CPU clock has been used to  measure the run-time. All the time is measured in seconds.
\par The Time 1 division is almost same for all the experiments. That is why we only mentioned it in the Table \ref{time} and did not plot. The Time 2 division was same for both the architecture InceptionresnetV2 and MobilenetV2, this is why only the MobilenetV2 curve is showing in the plot. 

\par In normal FL, let us assume on the average it takes $t$ amount of time for a user device to communicate with the central server. Therefore the total time will consist of: the central ask the user whether it is ready + the ready response from the device + the checkpoints and model sending time to the devices + training time + Dp reports of the weights and checkpoint generation + report sending; which is approximately $4t$ + training time + DP report generation time. Now, for our approach let there be $n$ users and $k$ cluster centers. Now for the $n-k$ users the time will be: the central ask the device + the device send ready + DP report generation of the data + DP report send to the cluster center. This is  approximately $2t$ + DP report generation + $t_c$, $t_c$ is the communication time from a user to the cluster center. This time is measured in the Time 1 and 2, most of the time which is less than 2 seconds. Therefore, approximately $(n-k)X2t$ number of communication channels will get cut from the original FL approach. Also, if we consider a scenario where most of the devices dropped and the normal FL system gets trained almost in a sequential manner, then, it becomes computationally expensive and the model quality becomes poor with respect to the central learning. The introduction of device clustering solves this issue, which can be seen from the last time graph for total computation time.   
\par Analysis of plots reveal that most of the time is taken by the Time 3 step or the training phase. Central learning takes the most total time and clustered and FL approach take almost similar or comparable time. If we sum up the Time 1 and Time 2 for every experiments the time overhead is almost 2 seconds. These summed up time measures the time limit of the data gathering at the cluster center. If any device gets lost, at this point, we can always look back to inspect but the training won't stop as there is data to provide, which solves the device drop problem. We can also see that more the number of weights, the Time 4 tends to get larger.
\begin{table*}
\caption{Time data for all four architectures}
\label{time}
\begin{tabular}{p{0.5cm}|p{0.55cm}|p{0.5cm}p{0.5cm}p{0.5cm}l|p{0.5cm}p{0.5cm}p{0.5cm}l|p{0.5cm}p{0.5cm}p{0.5cm}l|p{0.5cm}p{0.5cm}p{0.5cm}l}
\hline
\multicolumn{2}{c|}{} & \multicolumn{4}{c|}{MNIST-CNN} & \multicolumn{4}{c|}{CIFAR10\_CNN} & \multicolumn{4}{p{3cm}|}{Tiny-ImageNet-200-MobileNetV2} & \multicolumn{4}{p{3cm}}{Tiny-ImageNet-200-InceptionresnetV2} \\ \hline
Clus-ter     & t1       & t2     & t3     & t4   & tt    & t2     & t3     & t4    & tt      & t2    & t3      & t4   & tt      & t2     & t3       & t4     & tt        \\ \hline
2       & 0.083 & 0.183 & 22.14 & 1.11 & 23.516 & 0.69 & 18.79 & 0.81 & 20.76  & 1.083 & 545.91 & 2.13 & 548.813 & 1.083 & 600.21 & 4.92 & 600.98  \\
3       & 0.08  & 0.18  & 20.11 & 1.13 & 21.5   & 0.7  & 18.15 & 0.86 & 20.17  & 1.08  & 541.82 & 2.11 & 544.71  & 1.08  & 600.03 & 4.88 & 600.81  \\
4       & 0.079 & 0.179 & 19.24 & 1.12 & 20.618 & 0.72 & 16.82 & 0.87 & 18.85  & 1.079 & 539.71 & 2.03 & 542.539 & 1.079 & 599.93 & 4.86 & 600.73  \\
5       & 0.079 & 0.179 & 18.74 & 1.15 & 20.148 & 0.72 & 15.51 & 0.89 & 17.56  & 1.079 & 538.62 & 2.17 & 541.589 & 1.079 & 599.89 & 4.91 & 600.69  \\
6       & 0.079 & 0.179 & 18.41 & 0.96 & 19.628 & 0.72 & 14.62 & 0.82 & 16.6   & 1.079 & 538.3  & 2.15 & 541.249 & 1.079 & 599.8  & 4.82 & 600.6   \\
7       & 0.079 & 0.179 & 18.2  & 1.12 & 19.578 & 0.67 & 14.2  & 0.85 & 16.21  & 1.079 & 537.29 & 2.09 & 540.129 & 1.079 & 599.83 & 4.86 & 600.58  \\
8       & 0.079 & 0.179 & 17.76 & 0.93 & 18.948 & 0.67 & 13.88 & 0.88 & 15.92  & 1.079 & 537.18 & 2.1  & 540.029 & 1.079 & 599.53 & 4.82 & 600.28  \\
9       & 0.079 & 0.179 & 17.56 & 1.11 & 18.928 & 0.68 & 13.32 & 0.86 & 15.34  & 1.079 & 536.95 & 2.11 & 539.819 & 1.079 & 599.13 & 4.83 & 599.89  \\
10      & 0.079 & 0.179 & 17.62 & 1.15 & 19.028 & 0.69 & 12.97 & 0.87 & 15     & 1.079 & 535.89 & 2.13 & 538.789 & 1.079 & 598.99 & 4.84 & 599.76  \\
11      & 0.079 & 0.179 & 17.76 & 1.11 & 19.128 & 0.67 & 12.65 & 0.85 & 14.66  & 1.079 & 535.73 & 2.15 & 538.629 & 1.079 & 598.72 & 4.72 & 599.47  \\
12      & 0.079 & 0.179 & 17.67 & 0.97 & 18.898 & 0.68 & 11.67 & 0.89 & 13.72  & 1.079 & 535.51 & 2.11 & 538.379 & 1.079 & 598.5  & 4.77 & 599.26  \\
13      & 0.079 & 0.179 & 17.45 & 1.13 & 18.838 & 0.69 & 11.58 & 0.88 & 13.62  & 1.079 & 535.29 & 2.13 & 538.189 & 1.079 & 598.23 & 4.71 & 598.99  \\
14      & 0.079 & 0.179 & 17.23 & 0.99 & 18.478 & 0.68 & 11.43 & 0.81 & 13.4   & 1.079 & 534.22 & 2.12 & 537.099 & 1.079 & 598.02 & 4.78 & 598.78  \\
15      & 0.078 & 0.178 & 17.26 & 0.98 & 18.496 & 0.68 & 8.09  & 0.88 & 10.13  & 1.078 & 534.22 & 2.09 & 537.068 & 1.078 & 597.99 & 4.72 & 598.75  \\
16      & 0.078 & 0.178 & 17    & 1.19 & 18.446 & 0.69 & 8.03  & 0.82 & 10.01  & 1.078 & 534.12 & 2.08 & 536.968 & 1.078 & 597.91 & 4.72 & 598.68  \\
17      & 0.078 & 0.178 & 16.89 & 0.98 & 18.126 & 0.7  & 7.99  & 0.8  & 9.95   & 1.078 & 533.99 & 2.11 & 536.878 & 1.078 & 597.9  & 4.75 & 598.68  \\
18      & 0.078 & 0.178 & 16.35 & 0.95 & 17.556 & 0.72 & 8.06  & 0.91 & 10.13  & 1.078 & 533.95 & 2.16 & 536.908 & 1.078 & 597.88 & 4.76 & 598.68  \\
19      & 0.078 & 0.178 & 16.31 & 1.11 & 17.676 & 0.7  & 8.1   & 0.9  & 10.16  & 1.078 & 533.99 & 2.12 & 536.888 & 1.078 & 597.78 & 4.77 & 598.56  \\
20      & 0.076 & 0.176 & 15.14 & 0.94 & 16.332 & 0.7  & 7.98  & 0.82 & 9.95   & 1.076 & 533.63 & 2.14 & 536.546 & 1.076 & 597.41 & 4.72 & 598.2   \\ \hline
FL          &          &        & 5.37   & 0.95 & 6.32  &        & 9.27   & 0.88  & 10.15   &       & 533.3   & 2.11 & 535.41  &        & 597.22   & 4.82   & 602.04    \\ \hline
Cent-ral     &          &        &        &      & 41.14 &        &        &       & 169.69  &       &         &      & 3036.48 &        &          &        & 5182.44   \\ \hline

\end{tabular}
\end{table*}

\begin{table*}
\caption{Loss, AUC and f-score data for all four architectures}
\label{laf}
\begin{tabular}{p{0.65cm}|p{1cm}p{1cm}p{1cm}|p{1cm}p{1cm}p{1cm}|p{1cm}p{1cm}p{1cm}|p{1cm}p{1cm}p{1cm}}
\hline
        & \multicolumn{3}{|c}{MNIST-CNN}  & \multicolumn{3}{|c}{CIFAR-10-CNN} & \multicolumn{3}{|p{3cm}}{Tiny-ImageNet-200-MobinetV2}  & \multicolumn{3}{|p{3cm}}{Tiny-ImageNet-200-InceptionresnetV2} \\
        \hline
        Clus-ter & loss     & AUC      & f-score  & loss     & AUC      & f-score  & loss     & AUC      & f-score  & loss     & AUC      & f-score  \\
        \hline
2       & 0.029106 & 0.995    & 0.99912  & 0.647612 & 0.8878   & 0.9775   & 0.007813 & 0.744363 & 0.844363 & 0.013672 & 0.90248  & 0.97248  \\
3       & 0.029368 & 0.9955   & 0.9991   & 0.629345 & 0.88978  & 0.97856  & 0.015625 & 0.734335 & 0.834335 & 0.042969 & 0.858893 & 0.928893 \\
4       & 0.029309 & 0.9956   & 0.9992   & 0.63146  & 0.88705  & 0.977741 & 0.003906 & 0.748025 & 0.848025 & 0.078125 & 0.848004 & 0.918004 \\
5       & 0.03606  & 0.99494  & 0.99898  & 0.621674 & 0.890389 & 0.978078 & 0.021484 & 0.760447 & 0.860447 & 0.097656 & 0.841321 & 0.911321 \\
6       & 0.03266  & 0.995167 & 0.99903  & 0.664708 & 0.889167 & 0.9783   & 0.013672 & 0.760812 & 0.860812 & 0.111328 & 0.841079 & 0.911079 \\
7       & 0.032855 & 0.994834 & 0.998967 & 0.717057 & 0.88144  & 0.97678  & 0.011719 & 0.724809 & 0.824809 & 0.119141 & 0.833645 & 0.903645 \\
8       & 0.036057 & 0.99456  & 0.998911 & 0.680497 & 0.885278 & 0.977056 & 0.013672 & 0.771309 & 0.871309 & 0.134766 & 0.832965 & 0.902965 \\
9       & 0.03735  & 0.994611 & 0.998922 & 0.671125 & 0.883    & 0.9766   & 0.03125  & 0.746856 & 0.846856 & 0.15625  & 0.828764 & 0.898764 \\
10      & 0.034552 & 0.99483  & 0.998967 & 0.639487 & 0.89056  & 0.9781   & 0.021484 & 0.714381 & 0.814381 & 0.15625  & 0.818144 & 0.888144 \\
11      & 0.037673 & 0.99472  & 0.99894  & 0.690932 & 0.890278 & 0.97805  & 0.035156 & 0.782813 & 0.882813 & 0.173828 & 0.814217 & 0.884217 \\
12      & 0.33553  & 0.995    & 0.999    & 0.676812 & 0.886278 & 0.97725  & 0.029297 & 0.704479 & 0.804479 & 0.160156 & 0.813392 & 0.883392 \\
13      & 0.03434  & 0.9955   & 0.9991   & 0.639058 & 0.886167 & 0.97723  & 0.017578 & 0.703596 & 0.803596 & 0.191406 & 0.808335 & 0.878335 \\
14      & 0.032519 & 0.995056 & 0.999012 & 0.657092 & 0.88555  & 0.9777   & 0.033203 & 0.732946 & 0.832946 & 0.205078 & 0.804059 & 0.874059 \\
15      & 0.031592 & 0.9954   & 0.99089  & 0.657693 & 0.8878   & 0.97775  & 0.037109 & 0.799668 & 0.899668 & 0.207031 & 0.800149 & 0.870149 \\
16      & 0.03     & 0.995123 & 0.990118 & 0.6926   & 0.878    & 0.987655 & 0.048828 & 0.692272 & 0.792272 & 0.228516 & 0.800036 & 0.870036 \\
17      & 0.0295   & 0.991654 & 0.99014  & 0.626    & 0.891238 & 0.97654  & 0.033203 & 0.710671 & 0.810671 & 0.195313 & 0.789063 & 0.859063 \\
18      & 0.028    & 0.983195 & 0.999    & 0.576926 & 0.889988 & 0.971321 & 0.037109 & 0.727026 & 0.827026 & 0.208984 & 0.776445 & 0.846445 \\
19      & 0.031    & 0.995678 & 0.9989   & 0.66926  & 0.889123 & 0.977163 & 0.046875 & 0.741714 & 0.841714 & 0.248047 & 0.730246 & 0.800246 \\
20      & 0.032    & 0.995134 & 0.995689 & 0.576926 & 0.885643 & 0.956432 & 0.042969 & 0.74847  & 0.84847  & 0.238281 & 0.683008 & 0.753008 \\
\hline
FL      & 0.0367   & 0.9926   & 0.9811   & 0.68201  & 0.8822   & 0.9764   & 0.056641 & 0.793604 & 0.893604 & 0.265625 & 0.82765  & 0.89765  \\
\hline
Cent-ral & 0.0283   & 0.9956   & 0.99911  & 0.610001 & 0.892    & 0.9785   & 0.006344 & 0.80726  & 0.90726  & 0.165234 & 0.91825  & 0.98825  \\
      \hline        
\end{tabular}
\end{table*}

\section{Future Scope and Conclusion}\label{conclusion}

\par FLaPS is a novel distributed (federated) learning framework where the training is leveraged at a few cluster heads rather than all the devices individually. It builds on the foundations set by FL and DP built keeping user privacy in mind. Our approach in FLaPS gives a better privacy and utility guarantee by the introduction of DP in combination with FL in four simple steps. Scalability is added by introducing clustering approach at the beginning before the training. Following the lesser number of devices acting as cluster heads; robustness is improved with quicker re-initiation through smaller turn around time. Therefore in conclusion FLaPS builds a bridge between the central and FL with better robustness, privacy and scalability. In the future we would like to explore other variations in different phases of the system such as clustering, data aggregation, faster  as well as with different privacy settings. We would also like to test it on real mobile devices by setting up a test bed in our university.

\end{document}